\documentclass[journal]{IEEEtran}
\usepackage{amsmath,amsfonts}
\usepackage{algorithmic}
\usepackage{algorithm}
\usepackage{array}
\usepackage[caption=false,font=normalsize,labelfont=sf,textfont=sf]{subfig}
\usepackage{textcomp}
\usepackage{stfloats}
\usepackage{url}
\usepackage{verbatim}
\usepackage{graphicx}
\usepackage{cite}
\usepackage{multirow}
\usepackage{tcolorbox}
\usepackage{soul}
\usepackage{makecell}
\usepackage{booktabs}
\soulregister\cite7 
\soulregister\ref7  
\soulregister\label7 
\begin{document}

\title{Optimized Lattice-Structured Flexible EIT Sensor for Tactile Reconstruction and Classification}

\author{Huazhi Dong,\IEEEmembership{ Student Member, IEEE},
Sihao Teng,
Xu Han,
Xiaopeng Wu,\IEEEmembership{ Student Member, IEEE},
Francesco Giorgio-Serchi,\IEEEmembership{ Member, IEEE}, 
and Yunjie Yang,\IEEEmembership{ Senior Member, IEEE}
\thanks{Manuscript received March 31, 2025; revised March 31, 2025. This work was supported in part by the European Research Council Starting Grant under Grant no.101165927 (Project SELECT), the 2025 IEEE Instrumentation and Measurement Society Graduate Fellowship Award, and the WCSIM Post Graduate Scholarship Award. (Corresponding author: Yunjie Yang)}
\thanks{Huazhi Dong, Sihao Teng, Xu Han, Xiaopeng Wu and Yunjie Yang are with the SMART Group, Institute for Imaging, Data and Communications, School of Engineering, The University of Edinburgh, EH9 3BF Edinburgh, U.K. (e-mail: huazhi.dong@ed.ac.uk; sihao.teng@ed.ac.uk; s2143459@ed.ac.uk; xiaopeng.wu@ed.ac.uk; 
y.yang@ed.ac.uk).}
\thanks{Francesco Giorgio-Serchi is with the Institute for Integrated Micro and Nano Systems, School of Engineering, The University of Edinburgh, EH8 9YL Edinburgh, U.K. (e-mail: F.Giorgio-Serchi@ed.ac.uk).}}

\markboth{Journal of \LaTeX\ Class Files,~Vol.~14, No.~8, August~2021}%
{Shell \MakeLowercase{\textit{et al.}}: A Sample Article Using IEEEtran.cls for IEEE Journals}


\maketitle

\begin{abstract}
Flexible electrical impedance tomography (EIT) offers a promising alternative to traditional tactile sensing approaches, enabling low-cost, scalable, and deformable sensor designs. Here, we propose an optimized lattice-structured flexible EIT tactile sensor incorporating a hydrogel-based conductive layer, systematically designed through three-dimensional coupling field simulations to optimize structural parameters for enhanced sensitivity and robustness. By tuning the lattice channel width and conductive layer thickness, we achieve significant improvements in tactile reconstruction quality and classification performance. Experimental results demonstrate high-quality tactile reconstruction with correlation coefficients up to 0.9275, peak signal-to-noise ratios reaching 29.0303 dB, and structural similarity indexes up to 0.9660, while maintaining low relative errors down to 0.3798. Furthermore, the optimized sensor accurately classifies 12 distinct tactile stimuli with an accuracy reaching 99.6\%. These results highlight the potential of simulation-guided structural optimization for advancing flexible EIT-based tactile sensors toward practical applications in wearable systems, robotics, and human-machine interfaces. All data are publicly available in Edinburgh DataShare with the identifier https://doi.org/10.7488/ds/7982.
\end{abstract}


\section{Introduction}
\label{sec:introduction}
\IEEEPARstart{E}{Electronic} skin (e-skin) is an emerging sensing technology that employs flexible and stretchable materials integrated with embedded sensors to replicate the sensory functions of human skin\cite{booth2018omniskins}. By enabling robots to interact with and accurately perceive their surroundings, even in complex and unstructured environments \cite{li2022multifunctional}, e-skin represents a promising pathway toward embodied intelligence. Therefore, it has attracted significant attention for applications in human-machine interfaces (HMI) \cite{dai2024self}, health monitoring \cite{lai2023emerging}, and wearable devices \cite{lu2023wearable}. Typically, e-skin systems consist of a flexible substrate integrated with various sensors capable of detecting stimuli such as pressure, temperature, and humidity. Among these, tactile sensors are particularly crucial for identifying contact locations and quantifying applied forces. Existing tactile sensors are typically based on piezoresistive\cite{pohtongkam2021tactile}, capacitive\cite{niu2022ultrasensitive}, optical\cite{funk2024evetac}, and magnetic\cite{hu2022wireless} mechanisms. However, these approaches generally utilize arrays of discrete sensing units \cite{wu2024bimodal} that require intricate wiring and complex structures, leading to higher production costs, limited flexibility and scalability. Scaling such sensors to larger areas presents challenges such as reduced spatial resolution, higher costs and hardware complexity \cite{wang2023tactile}, thereby restricting their practical deployment in real-world applications. 

Electrical impedance tomography (EIT) has emerged as a compelling alternative for tactile sensing, addressing many limitations of conventional array-based designs \cite{van2020large}. EIT reconstructs the conductivity distribution within the Region of Interest (ROI) by analysing voltage readouts from sparsely distributed boundary electrodes. This enables the detection of contact locations and applied forces without the need for dense sensor arrays. Its “one-piece” architecture with sparse boundary electrodes offers advantages in terms of scalability, mechanical compliance, and manufacturing simplicity. The integration of flexible materials such as ionic liquids\cite{chen2024correcting}, hydrogels\cite{dong2024learning}, and elastic films\cite{wu2022new} has further enhanced the stretchability, sensitivity, and self-healing capabilities. Recent developments have demonstrated EIT-based tactile sensors can effectively track dynamic pressure changes\cite{kim2024extremely}. Their capability in capturing temporal variations makes them particularly suitable for applications requiring continuous and responsive tactile sensing.

Conventional EIT-based tactile sensors usually adopt uniformly distributed conductive pathways, resulting in only small potential changes at the boundary electrodes when localised pressure is applied. This inherent characteristic limits the sensitivity of EIT-based tactile sensing systems. Recent research has explored novel structures to improve sensitivity, such as porous\cite{chen2022large} and multi-layer structures\cite{park2021deep}. However, these approaches often face manufacturing challenges that constrain scalability and durability. Furthermore, the lattice structure facilitates higher current density, enabling more pronounced and rapid conductivity changes under press deformation, thereby significantly improving the responsiveness of the sensing system \cite{jamshidi2024design}. However, current lattice-structured EIT tactile sensors face two key limitations\. First, critical lattice parameters, such as channel width and conductive layer thickness, are often selected empirically, without systematic optimization. Second, the influence of lattice structure on the EIT signal response has not been quantitatively analyzed or benchmarked against non-lattice configurations. These gaps limit the development of high-sensitivity, robust, and reproducible tactile sensors.  

In this work, to resolve the above limitations, we introduce an optimized lattice-structured flexible EIT tactile sensor, where the lattice parameters are systematically refined through three-dimensional coupling field simulations (3D-CFS). This simulation-guided design approach enables a balance of sensitivity, spatial resolution and mechanical robustness. The key contributions of this work are as follows.
\begin{itemize}
    \item We develop a flexible lattice-patterned EIT-based tactile sensor fabricated with hydrogel and silicone, whose lattice design parameters (specifically channel width and conductive layer thickness) are optimized through 3D-CFS to maximize sensing performance.
    \item We validate the optimized sensor’s capabilities with real-world experiments, demonstrating significantly enhanced sensitivity enabled by the optimized lattice-patterned structure. 
    \item We demonstrate the optimized sensor’s effectiveness in tactile reconstruction and classification tasks. The sensor achieves superior reconstruction quality and successfully recognizes 12 distinct static and dynamic touch patterns, outperforming comparable EIT-based tactile sensing studies \cite{park2022biomimetic,yang2023robotic,yang2024body}.
\end{itemize}

\section{Principle of EIT-based tactile sensing}
\begin{figure}[t]
\centerline{\includegraphics[width=\columnwidth]{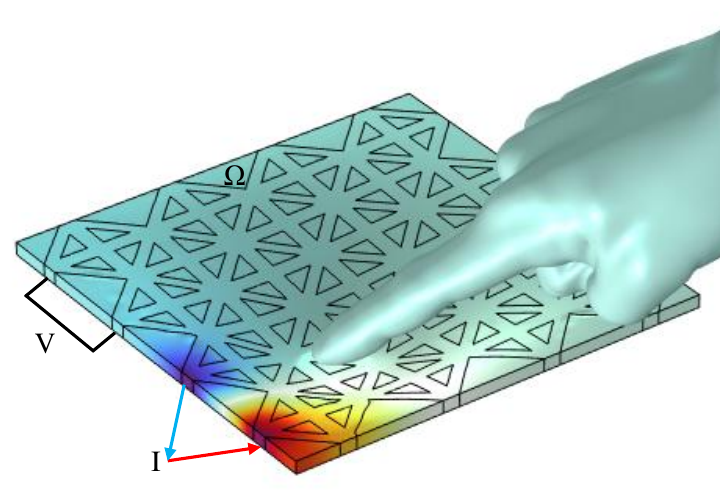}}
\caption{Schematic of lattice-structured flexible EIT sensor for tactile sensing.}
\label{fig-EitTheory}
\end{figure}
The principle of EIT-based tactile sensing is illustrated in Fig. \ref{fig-EitTheory}. When localized pressure is applied to a sample substrate, it induces changes in conductivity within the touched regions, which in turn alter the electric potentials measured at the boundary electrodes. The objective of EIT-reconstruction in tactile sensing is to estimate the pressure-induced conductivity changes $ \boldsymbol{\sigma}$ in the sensing region $ \boldsymbol{\Omega} $ based on the voltage measurements $ \boldsymbol{V} $. EIT involves solving both a forward and an inverse problem. The forward problem predicts surface potentials for a given conductivity and applied current. This relationship is governed by:
\begin{equation}
    \nabla \cdot \left [{ \sigma \left ({x,y }\right)\nabla u\left ({x,y }\right) }\right]=0,\quad \left ({x,y }\right)\in \boldsymbol{\Omega} \tag{1}
    \label{eq1}
\end{equation}
where $\sigma \left ({x,y }\right)$ represents the conductivity at $\left ({x,y }\right)$, and $u \left ({x,y }\right)$ is the potential distribution. This can be linearized as follows:
\begin{equation}
    \Delta \boldsymbol{V}=\boldsymbol{J}\Delta\boldsymbol{\sigma}\tag{2}
    \label{eq2}
\end{equation}
where $\Delta$ denotes discrete change and $\boldsymbol{J}$ is the Jacobian matrix. The EIT inverse problem involves estimating $ \Delta \boldsymbol{\sigma} $ from $\Delta \boldsymbol{V} $, which is typically formulated as a regularized optimization problem:
\begin{equation}
    \displaystyle \arg\min_{\Delta\boldsymbol{\sigma}}\frac{1}{2}\Vert \boldsymbol{J}\Delta\boldsymbol{\sigma}-\Delta \boldsymbol{V}\Vert_{2}^{2}+\lambda R(\Delta\boldsymbol{\sigma})\tag{3}
    \label{eq3}
\end{equation}
where $R$ is the regularization term incorporating prior knowledge, and $\lambda>0$ is the regularization factor that balances data fidelity and regularization. 

In contrast, learning-based approaches for solving the EIT inverse problem aim to establish an inverse mapping operator, denoted as $\mathcal{F}$, using data-driven methods:
\begin{equation}
    \Delta \hat{\sigma}=\mathcal{F}(\Delta V)\tag{4}
    \label{eq4}
\end{equation}

\section{Simulation-Guided Structural Optimization and Experimental Validation}
\begin{figure}[!t]
\centering
\includegraphics[width=\columnwidth]{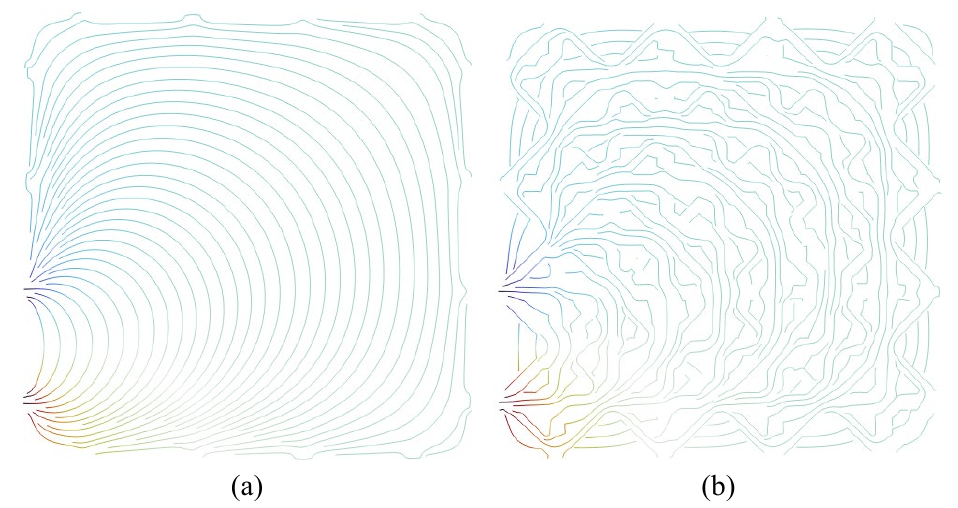}
\caption{Electric field distribution for (a) uniform conductivity layer and (b) lattice-patterned conductivity layer.}
\label{fig-Electricfield}
\end{figure}
\subsection{Physical Mechanism}
Compared to sensors with a uniformly conductive medium, lattice-pattern EIT tactile sensors exhibit more pronounced voltage changes. This enhanced sensitivity primarily arises from two interrelated factors: increased local current density and modified electric field distributions \cite{jamshidi2023eit}.

In a lattice-patterned sensor, the conducting material is restricted to predetermined channels arranged in a lattice grid pattern while the rest of the area is non-conductive. This design forces current to flow primarily through a limited number of channels, thereby increasing the local current density. When external pressure is applied, the geometry of these channels changes, altering their cross-sectional area and shape, thereby yielding significant local resistance changes.

The electric field lines in traditional EIT sensors are distributed relatively uniformly throughout the conducting medium (Fig. \ref{fig-Electricfield}a). In contrast, lattice-patterned structures induce sharper electric field gradients within the conductive channels (Fig. \ref{fig-Electricfield}b).  Upon external pressure, the deformation of the channels further intensifies these localized field gradients, amplifying the sensor’s impedance response.

\subsection{3D-CFS}
\begin{figure}[!t]
\centering
\includegraphics[width=\columnwidth]{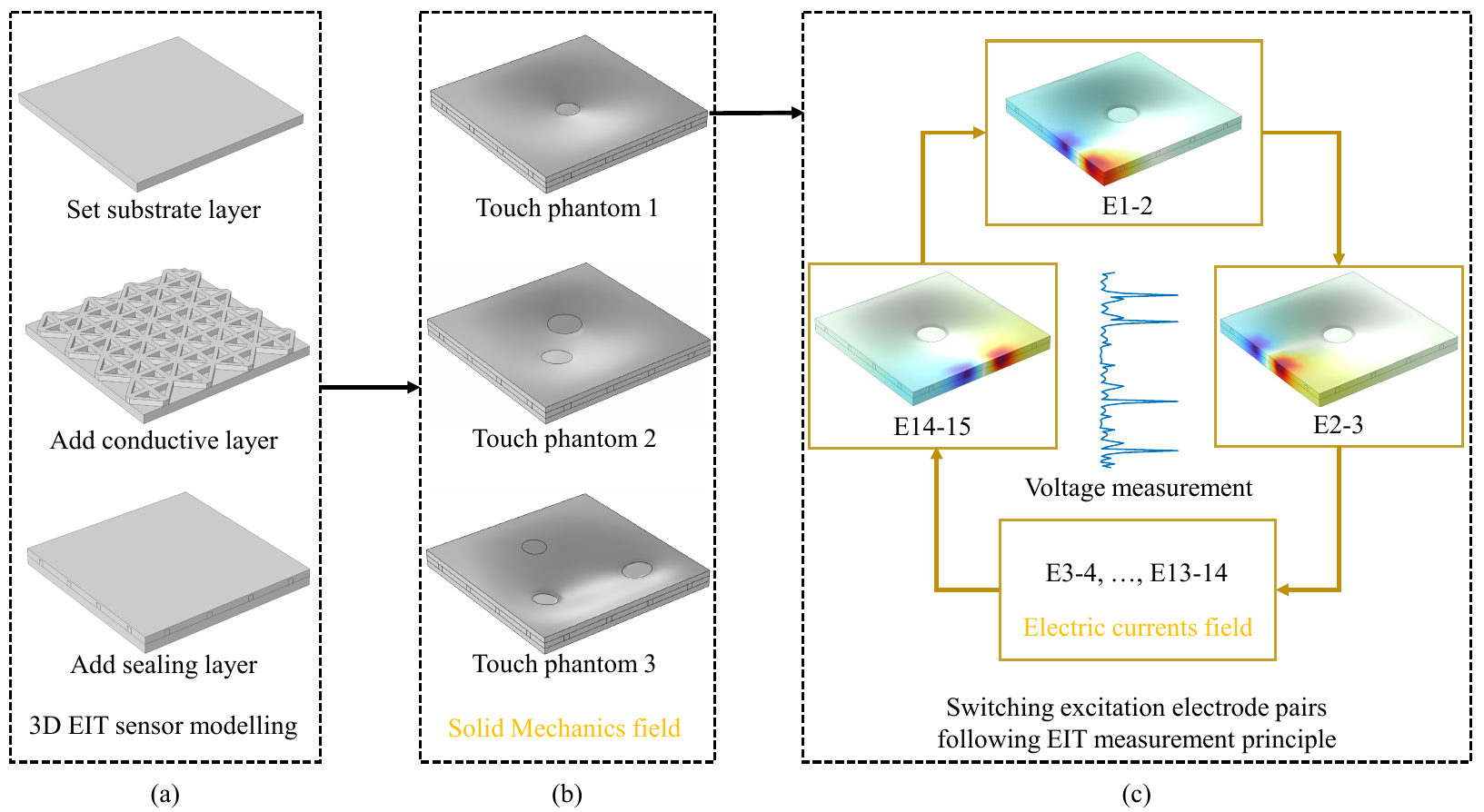}
\caption{Framework of 3D-CFS. (a) 3D EIT tactile sensor modelling. (b) Solid mechanics field. Touch phantom 1 is the example for one touch point, touch phantom 2 is the example for two touch points, and touch phantom 3 is the example for three touch points. (c) Electric current field. E1, ..., E15 represent the index of electrodes.}
\label{fig-cfs}
\end{figure}
\begin{figure}[!t]
\centering
\includegraphics[width=\columnwidth]{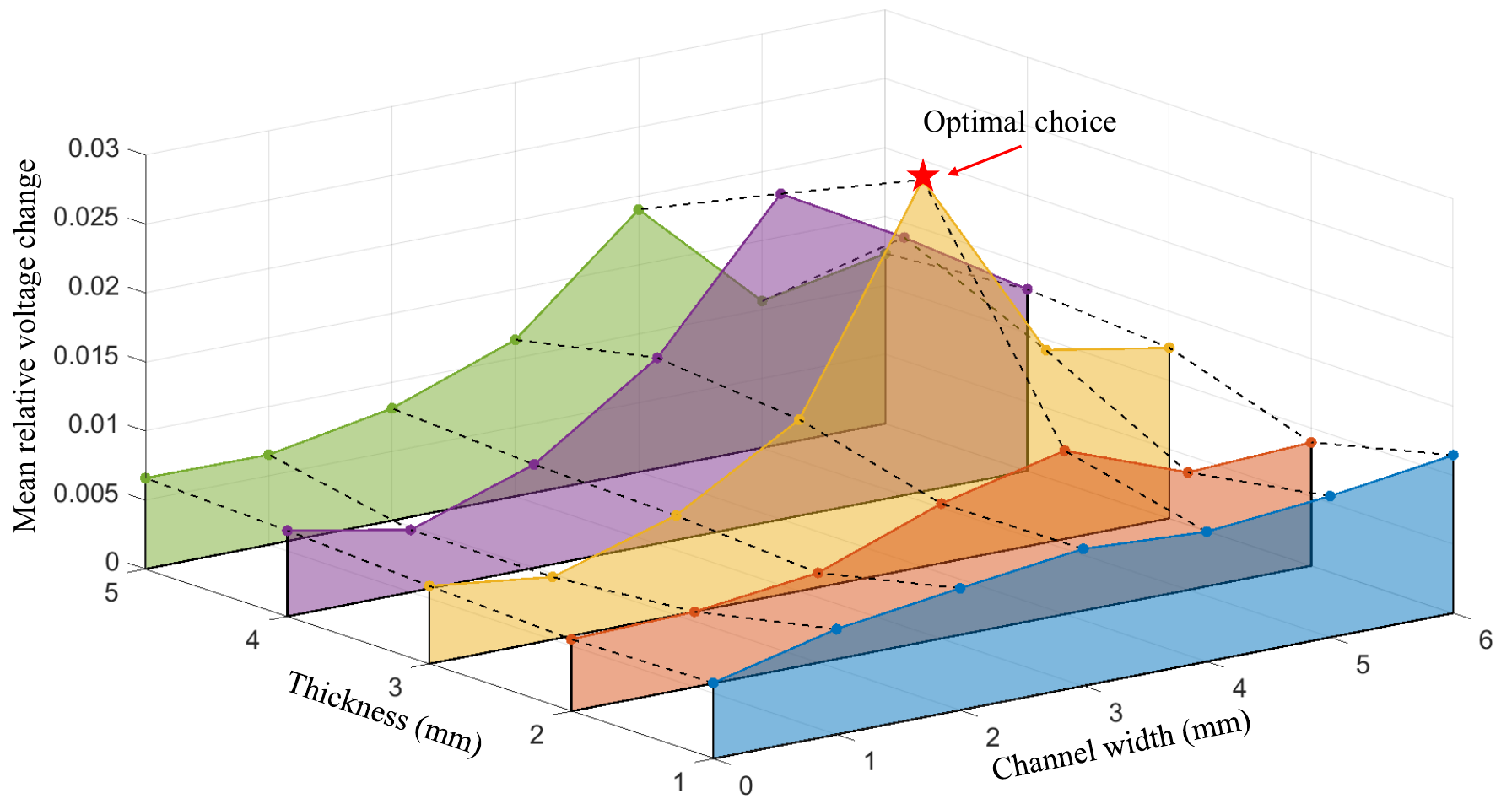}
\caption{Mean relative voltage changes for 300 random touch phantoms against various lattice parameters.}
\label{fig-CFSResult}
\end{figure}
\begin{figure}[!t]
\centering
\includegraphics[width=\columnwidth]{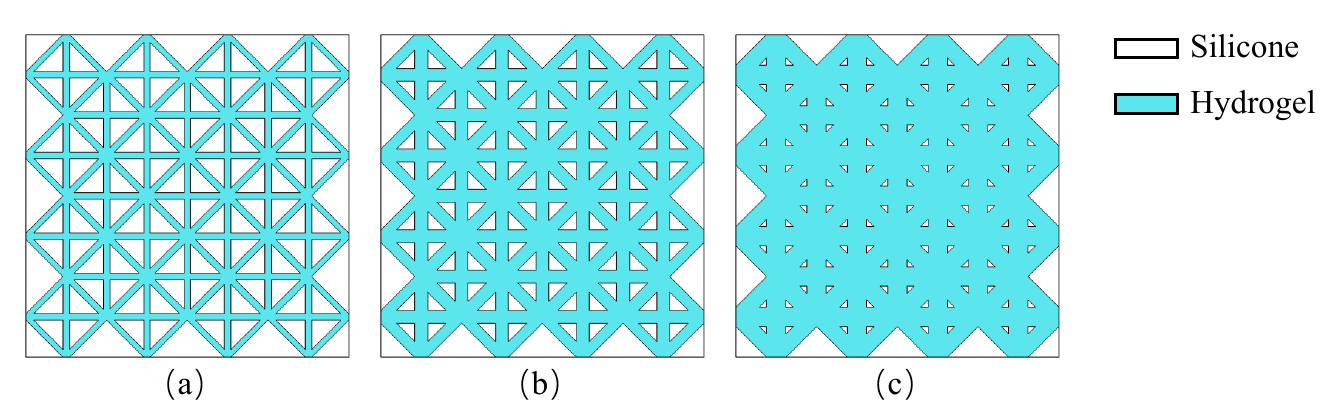}
\caption{Visualization of lattice pattern geometries under different conductive channel widths. 
(a) $w = 2$ mm; (b) $w = 4$ mm (optimized); (c) $w = 6$ mm. 
Blue regions represent the conductive hydrogel channels, while white regions represent the silicone area.}
\label{fig-LatticeView}
\end{figure}
We conducted 3D-CFS to determine the optimal layer thickness and lattice channel width for maximizing sensor sensitivity. The 3D-CFS framework combines a solid mechanics field to model the deformation caused by external pressure under different touch patterns, with an electric current field model that simulates the injected current through electrode pairs positioned along the model boundaries. The adjacent-drive adjacent-measurement excitation protocol was employed to obtain 104 voltage measurements for each touch pattern.

As shown in Fig. \ref{fig-cfs}a, the simulated sensor model is designed to be 100 × 100 × 9 mm and consists of three layers and 16 evenly distributed boundary electrodes. The substrate and sealing layers are non-conductive, and the sealing layer has a fixed thickness of 2 mm. The thickness of the substrate layer is changed with the intermediate conductive lattice layer to maintain a constant total model thickness of 9 mm. To evaluate and optimise sensor sensitivity, six different lattice channel widths and five different conductivity layer thicknesses were simulated to assess the influence of lattice channel widths and conductive layer thickness on sensor sensitivity.

As shown in Fig. \ref{fig-cfs}b, three different touch conditions are applied by varying the number of touch points from one to three. In each condition, 100 sets of touch phantoms are randomly generated, with press depths from 1 mm to 5 mm and touch radius from 6.25 mm to 8.75 mm.

To determine the optimal lattice configuration, the average relative voltage change $V^{\text{rel}}$ was introduced to quantify the sensor's electrical response to external tactile stimuli.
\begin{equation}
    V^{\text{rel}} = \frac{1}{N} \sum_{i=1}^{N} \frac{|V_i^{\text{touch}} - V_i^{\text{ref}}|}{|V_i^{\text{ref}}|} \tag{5}
    \label{eq5}
\end{equation}
where $N$ is the total number of measurement channels; $V_i^{\text{touch}}$ represents the voltage measured at the i-th channel under touch conditions, and $V_i^{\text{ref}}$ is the corresponding voltage under the reference (no-touch) condition. 

By averaging the normalized voltage differences across all channels, this metric captures the overall sensitivity of the sensor configuration to pressure-induced conductivity changes.

Different lattice parameters were calculated, with the conductive layer thickness from 1 mm to 5 mm and the channel width from 0 mm (no lattice pattern) to 6 mm. The result is as shown in Fig. \ref{fig-CFSResult}, each average value represents the average relative voltage change from 300 random touch phantoms across three touch conditions. We defined the optimal structural parameters as a conductive layer thickness of $t = 3$~mm and a conductive channel width of $w = 4$~mm. These values were selected because they maximized the average relative voltage change $V^{\text{rel}}$ under touch indentation, while the other fabricated variants are considered “non-optimized” for comparison. To provide a more intuitive understanding of the geometric differences, we visualized the lattice patterns corresponding to the optimized configuration (Fig.~\ref{fig-LatticeView}b, $w = 4$~mm) and two representative non-optimized variants (Fig.~\ref{fig-LatticeView}a, $w = 2$~mm; Fig.~\ref{fig-LatticeView}c, $w = 6$~mm).
\subsection{Sensor Fabrication}
\begin{figure}[!t]
\centering
\includegraphics[width=\columnwidth]{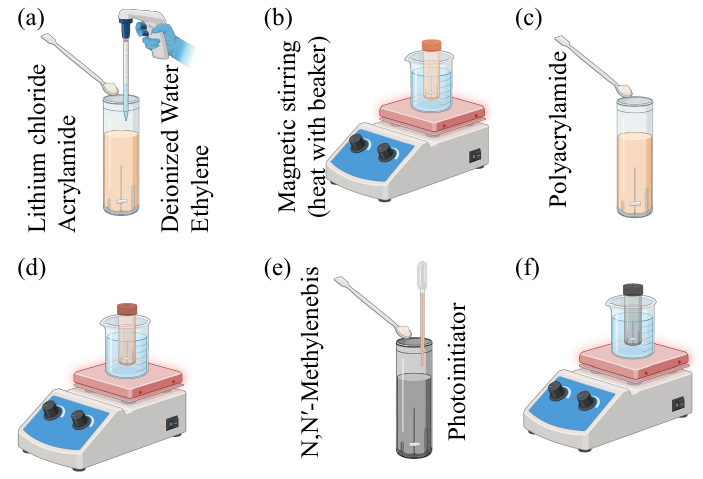}
\caption{Hydrogel fabrication process. (a) The solution contained 21.5 weight percent (wt\%) lithium chloride (Sigma-Aldrich), 8 wt\% acrylamide (Sigma-Aldrich), 32.27 wt\% deionised water and 37 wt\% ethylene glycol (Sigma-Aldrich) was prepared. Ethylene glycol maintains hydrogel pliability across a wide temperature range, essential for stable sensor performance. (b) The solution was stirred at 60 ${ }^{\circ} \mathrm{C}$ and 700 rpm for two hours to achieve uniform mixing and dissolution. (c) To initiate a polyacrylamide-based network, 1 wt\% polyacrylamide (Sigma-Aldrich) was introduced, providing mechanical stability and enhancing sensor durability. (d) The solution was stirred overnight at 60 ${ }^{\circ} \mathrm{C}$ and 700 rpm to ensure complete dissolution and formation of a uniform polymer network. (e) A crosslinking agent, 0.15 wt\%  N,N’-methylenebisacrylamide (Sigma-Aldrich), was added to control the mechanical strength of the gel. Additionally, 0.08 wt\%  2-hydroxy-2-methylpropiophenone (Sigma-Aldrich), a photoinitiator, was added to enable UV curing of the hydrogel. (f) The mixture was shielded from light and stirred for an additional two hours at 60 ${ }^{\circ} \mathrm{C}$ and 700 rpm to ensure complete homogenization.}
\label{fig-hydrogel}
\end{figure}
\begin{figure*}[!ht]
\centering
\includegraphics[width=\textwidth]{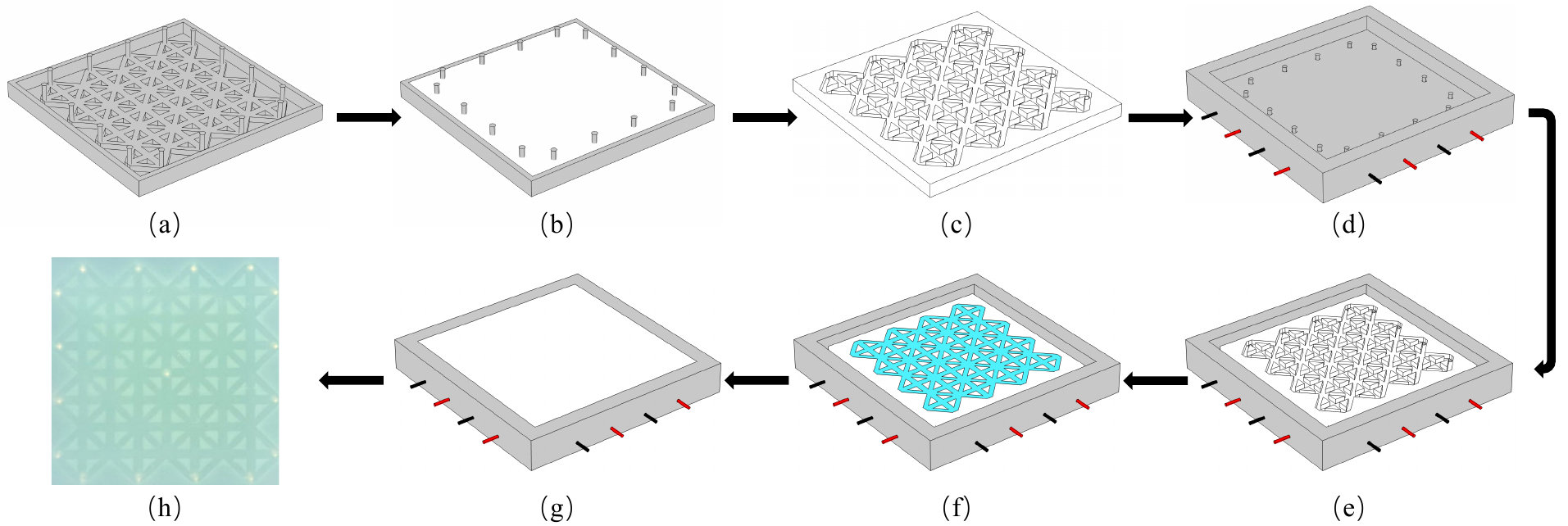}
\caption{Fabrication process of the lattice-patterned tactile sensor. (a) A 3D-printed mould A was prepared to define the sensor structure. (b) A pre-gel mixture of silicone was poured into the mould. This mixture was prepared by combining parts A and B of Ecoflex 00-30 (Smooth-On Inc.) in a 1:1 ratio. (c) After allowing the silicone to cure for 4 hours at room temperature, the substrate layer was released from the mould. (d) A 3D-printed mould B, containing designated regions for the sensing area, cable channels, and electrode holes, was prepared. Cables and electrodes were installed within the mould. (e)The released substrate layer was assembled with the sealed 3D-printed mould B, and the gap was sealed with silicone. (f) The hydrogel precursor was poured into the designated sensing area and cured under UV light (365 nm) for two hours. (g) A final layer of silicone pre-gel was applied over the entire sensor to form a sealing layer, protecting it from environmental interference. (h) The fabricated lattice-patterned tactile sensor.}
\label{fig-sensor}
\end{figure*}
Based on the optimal design parameters identified through the 3D-CFS simulations, a conductive layer thickness of 3 mm and a lattice channel width of 4 mm, we proceeded to fabricate the lattice-structured EIT tactile sensor accordingly. The sensing layer was fabricated from hydrogel, and the hydrogel precursor synthesis procedure was shown in Fig. \ref{fig-hydrogel}. As shown in Fig. \ref{fig-sensor}, the hydrogel and silicone layers were integrated to fabricate the complete tactile sensor (Fig. \ref{fig-sensor}h).

To enhance durability and maintain long-term functionality, two mitigation strategies against water evaporation were adopted. First, lithium chloride was selected as the ionic conductor in the hydrogel precursor formulation due to its hygroscopic properties, which effectively suppress dehydration while maintaining high ionic conductivity. Second, the entire hydrogel layer was encapsulated with a stretchable silicone elastomer to serve as a physical barrier against moisture loss \cite{zhu20203d}.

\subsection{Real-world Sensitivity Characterization and Validation}
\begin{figure*}[!ht]
\centering
\includegraphics[width=\textwidth]{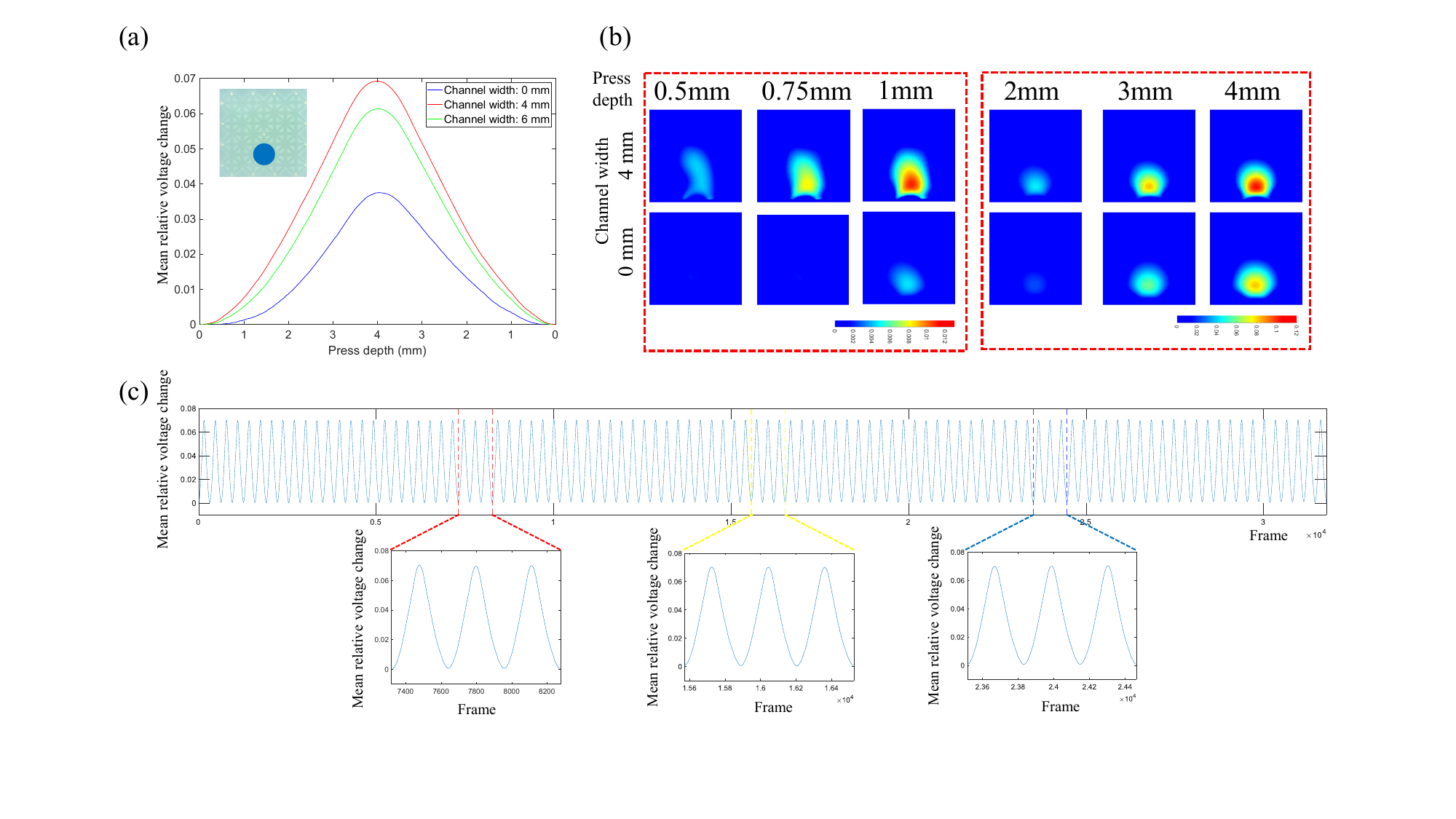}
\caption{Sensitivity characterization of different sensors. (a) Mean relative voltage change of three sensors. (b) Tactile reconstruction using Tikhonov Regularization \cite{Vauhkonen1998} for lattice-patterned and uniform sensors. (c) Sensor cycling test of the lattice-patterned sensor for 100 pressing cycles.}
\label{fig-SensorPress}
\end{figure*}
To verify the optimized sensor parameters identified through CFS, we fabricated three sensors using the same procedures (refer to Fig. \ref{fig-sensor}) and a consistent conductive layer thickness ($t = 3$~mm): two lattice-patterned sensors with different lattice parameters ($w = 4$ and $6$~mm) and one without the lattice structure for comparison. These sensors were subjected to a continuously increasing localized pressure with strains ranging from 0 mm to 4 mm, as illustrated in Fig. \ref{fig-SensorPress}a. The pressure was applied through a circular indenter with a radius of 8.75 mm, simulating the typical contact area of a human fingertip. The pressing location, marked by the blue dot in the inset of Fig. \ref{fig-SensorPress}a, remained consistent across all measurements. The results demonstrate that the sensor with optimized lattice parameters ($w = 4$~mm) consistently shows superior sensitivity compared to the non-lattice structured sensor and lattice-structure sensor with non-optimal parameters ($w = 6$~mm), aligning well with the simulation results. Specifically, at the press depth of 4 mm, the average relative voltage change $V^{\text{rel}}$ for the sensor with the optimized lattice ($w = 4$~mm) reaches approximately 0.0692, compared to ~0.0613 for the non-optimized lattice sensor ($w = 6$~mm) and only ~0.0375 for the non-lattice sensor. This corresponds to an 84.5\% improvement over the non-lattice structure and a 12.9\% enhancement compared to the non-optimized parameters ($w = 6$~mm), clearly confirming the effectiveness of the optimized lattice pattern in enhancing sensitivity.

Additionally, tactile reconstruction results comparing the optimized lattice-patterned sensor and the sensor without the lattice channel are illustrated in Fig. \ref{fig-SensorPress}b. The reconstructed images show that the optimized lattice-patterned sensor successfully reconstructs tactile information even at lower compressive strains, whereas the sensor without the lattice structure fails to identify tactile input at these shallow press depths. Furthermore, as the compressive strain increases, the lattice-patterned sensor consistently demonstrates superior reconstructions compared to the sensor without the lattice structure. These results further validate that the lattice-patterned sensor possesses higher sensitivity and enhanced tactile reconstruction performance.

\subsection{Durability Evaluation via Cycling Test}
To further evaluate the durability and repeatability of the optimized lattice-patterned sensor, we performed a cycling test involving 100 pressing cycles with compressive strains from 0 mm to 4 mm. The contact area and location for the pressure keep the same with Fig. \ref{fig-SensorPress}a. The results show that the sensor maintains stable output throughout the cycles, with consistent trends in signal changes (Fig. \ref{fig-SensorPress}c). These results confirm the sensor's robustness and suitability for applications requiring long-term, repeated tactile interactions.

\section{Tactile reconstruction: Spatial Resolution Evaluation}
\begin{figure*}[!ht]
\centering
\includegraphics[width=\textwidth]{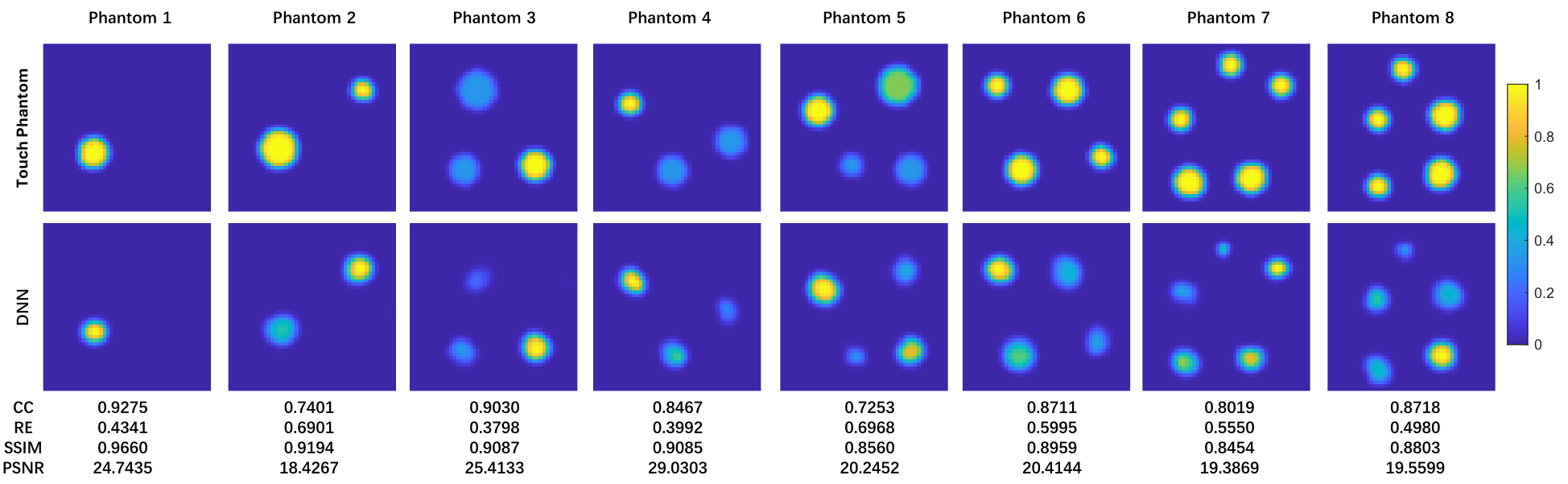}
\caption{Tactile reconstruction using experiment data. Images are normalized.}
\label{fig-recon1}
\end{figure*}
\begin{figure*}[!ht]
\centering
\includegraphics[width=\textwidth]{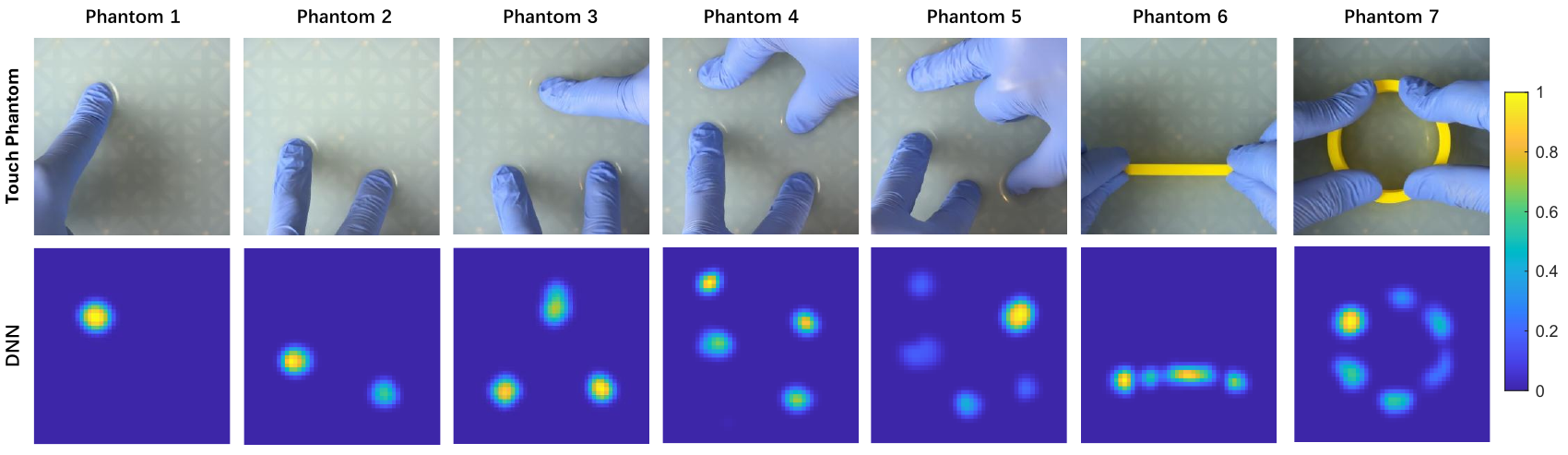}
\caption{Tactile reconstruction of unseen real-world complex touches. Images are normalized.}
\label{fig-recon2}
\end{figure*}
\subsection{Dataset and Network Training}
To validate the sensor’s tactile reconstruction performance, we adopted the deep neural network (DNN) architecture in \cite{dong2025data}. This model employs the Multilayer Perceptron (MLP) to extract global signal characteristics of voltage measurements, followed by a Convolution Neural Network (CNN) that extracts fine-grained local patterns. 

To generate the dataset for the lattice-structured EIT sensor, we constructed a finite element model (FEM) using COMSOL Multiphysics and Matlab. We divide the sensing region with a size of 100 $\times$ 100 mm${^2}$ into 2304 elements (48 $\times$ 48) and set the background conductivity $\boldsymbol{\sigma_0}$ to 0.00312 S/m. Tactile interactions were simulated using circular regions with diameters ranging from 7.5 mm to 27.5 mm, varying in both location and intensity. The conductivity within these touch regions is randomly assigned values between $0.05 \times \sigma_0$ and $2 \times \sigma_0$. The complete dataset comprises 50,000 samples, divided into five subsets containing 10,000 samples each. These subsets correspond to different numbers of touch regions, ranging from 1 to 5 touch areas per sample. Each subgroup is further partitioned into training, validation, and testing sets using a 7:2:1 ratio. To enrich the training dataset by supplementing uncollected location information, we used a data augmentation approach to convert a single EIT measurement into 8 distinct readouts \cite{dong2025data}. To further improve generalisation, additive Gaussian noise with Signal-to-Noise Ratios (SNRs) of 35, 40, 45, 50, 55, 60 and 65 dB was added to the seven augmented measurements, respectively. Finally, this process yielded a total of 280,000 samples for training, 10,000 samples for validation, and 5,000 samples for testing.

For DNN training, we retained the parameter settings and optimization strategy reported in \cite{dong2025data}. To match the target resolution of 48 $\times$ 48 in this study, an upsampling layer was appended to rescale the original 64 $\times$ 64 output.

\subsection{Reconstruction Results and Discussion}
As shown in Fig. \ref{fig-recon1}, from Phantom 1 to Phantom 8, we applied 1 to 5 contacts at different positions using a linear robotic platform (DLE-RG-0001). The reconstruction results demonstrate that the sensor with the DNN model successfully reconstructs various touch patterns with good accuracy. To quantitatively assess reconstruction quality, we evaluate the results using correlation coefficients (CC), peak signal-to-noise ratio (PSNR), relative image error (RE), and structural similarity indexes (SSIM) \cite{wang2004image,yu2024high}. Across all phantoms, the model achieves strong performance, with CC values reaching up to 0.9275, RE values as low as 0.3798, SSIM values up to 0.9660, and PSNR values reaching 29.0303 dB. These results indicate that the proposed sensor and model combination delivers high-fidelity reconstructions with strong consistency and visual clarity.

Beyond the above benchmark scenarios, we further validated the system’s performance on real-world measurements involving more complex and natural touch patterns. These include rectangular contact regions and annular touch profiles. Neither of them is included in the training dataset, so these cases serve as a strong test of the model's generalization capability. As shown in Fig. \ref{fig-recon2}, the DNN model successfully reconstructs these diverse tactile configurations with clear spatial patterns and consistent localization, demonstrating that the learned representation is robust to previously unseen contact geometries and pressure distributions. For instance, in Phantom 6 and Phantom 7, which correspond to rectangular and annular touch profiles, respectively, the model approximates these shapes using combinations of circular blobs. This behavior arises from the fact that the training data only included circular touch patterns; nonetheless, the model is still able to capture the overall geometry and spatial distribution.

This generalization stems from the training dataset incorporates a wide range of touch sizes, intensities, and noise perturbations, enabling the model to learn transferable spatial priors. Supplementary Video 1 further demonstrates the model’s real-time tactile reconstruction capabilities in these challenging scenarios, reinforcing its suitability for practical, interactive applications with arbitrary and dynamic touch inputs.

\section{Human–Machine Interaction Demonstration}
\begin{figure}[t]
\centering
\includegraphics[width=\columnwidth]{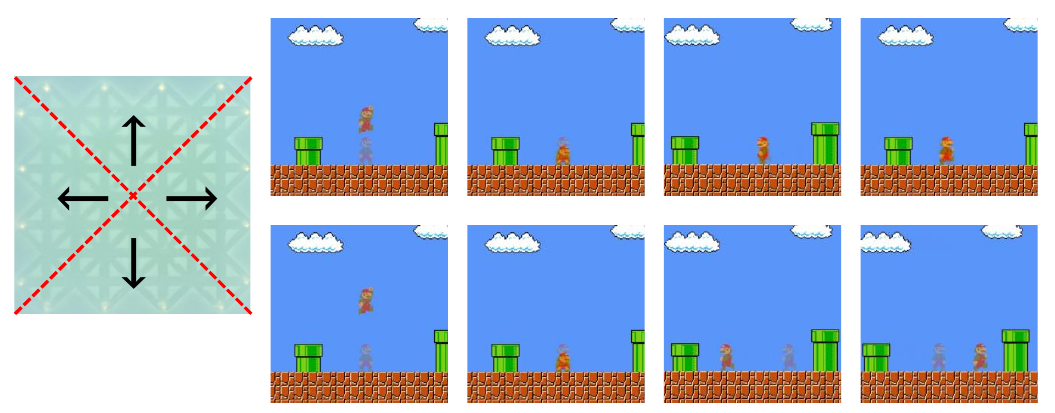}
\caption{HMI application: controlling the Super Mario Bros Game.}
\label{fig-HMI}
\end{figure}
To demonstrate the practical utility of the proposed tactile sensor, we implemented it as a human–machine interface (HMI) for controlling a virtual character in the Super Mario Bros game, as shown in Fig. \ref{fig-HMI}. The left column shows the physical tactile contact on the sensor, and the right column describes the actions generated in the virtual environment, illustrating the character's response to each tactile input. When the user presses the sensor in different locations, the corresponding action is performed in the game, such as advancing or jumping. When the user presses the sensor for a different time, it will also be recognised as a distinct motion amplitude, such as low-altitude jumping and high-altitude jumping. Supplementary Video 2 showcases real-time interactions with the Super Mario Bros Game. These results confirm the feasibility of using our tactile sensors for real-time, intuitive control in virtual environments, underscoring their ability to capture and transform subtle tactile inputs for multi-functional HMI applications.

\section{Tactile Classification: Temporal Trajectory Recognition}
\subsection{Dataset and Network Training}
\begin{figure}[t]
\centerline{\includegraphics[scale=0.5]{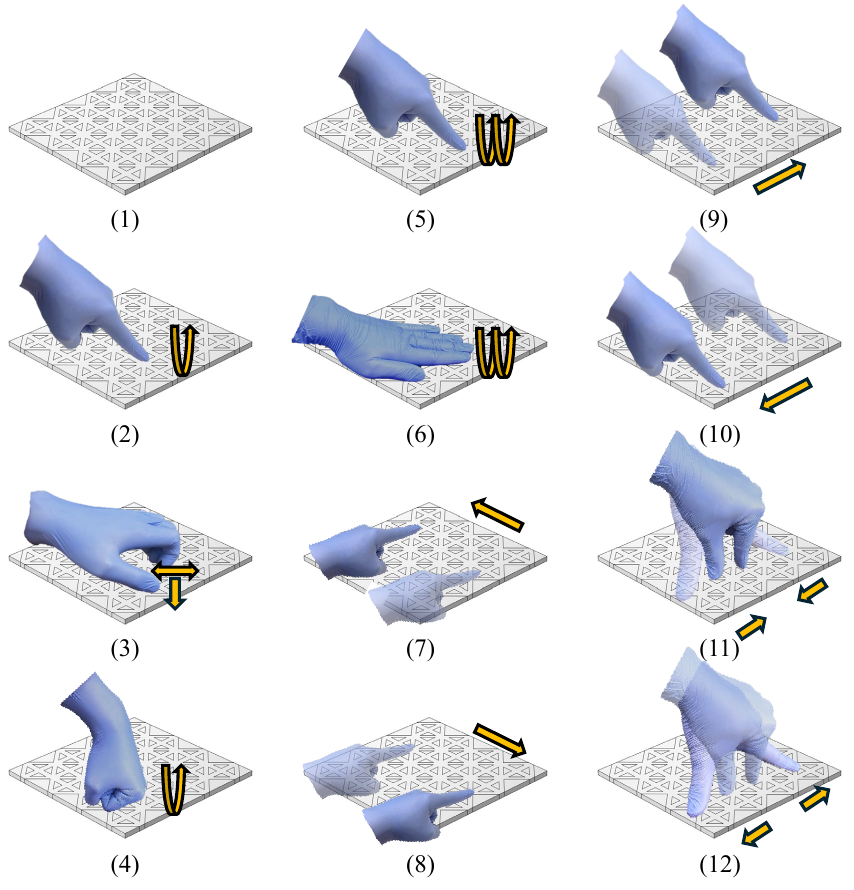}}
\caption{12 classes of tactile stimulation.}
\label{fig-12class}
\end{figure}

To validate the superior spatio-temporal characteristics of our sensor, we conducted continuous motion detection experiments using various tactile interactions. Specifically, we selected tactile gestures with different movement trajectories as reference actions to assess the system's performance. As shown in Fig. \ref{fig-12class}, the dataset includes 12 distinct interaction types such as no contact, finger press, four-finger scratch, fist press, finger double tap, palm pat, finger swipe up, finger swipe down, finger swipe left, finger swipe right, zoom-in and zoom-out. Each interaction type comprises 1000 samples, each containing 15 frames of voltage measurements. Then, we divided the 12,000 signals into a training set and a validation set in an 8:2 ratio. To fully evaluate the generalisation and robustness, 1,200 new signals were collected to be the test dataset, with each interaction type comprising 100 samples.

To validate the effectiveness and generalizability of the input data, we implemented two types of models: a CNN-based architecture adapted from \cite{park2022biomimetic,yang2023robotic,yang2024body}, and a Long Short-Term Memory (LSTM) model. By comparing the performance of both models, we aim to assess the suitability of our data for both spatial and temporal learning approaches. The model parameters are listed in Table \ref{tab-trainparams}.

\begin{table}[t]
    \centering
    \caption{Model parameters}
    \setlength{\tabcolsep}{3pt} 
    \renewcommand{\arraystretch}{1.2} 
    \resizebox{\columnwidth}{!}{ 
    \begin{tabular}{lcc} 
    \toprule
    Parameter & CNN & LSTM\\ 
    \midrule
    Optimizer type & Adam & Adam \\
    Base learning rate & 1.0e-4 & 1.0e-4 \\
    Total epoch & 200 & 200 \\
    The epoch of Best Model & 60 & 72 \\
    Loss function & Cross Entropy Loss & Cross Entropy Loss\\
    Batch size  & 256 & 256\\
    Input dimension & 1$\times$15$\times$104 & 15$\times$104\\
    Output dimension & 12 & 12\\
    Hardware & RTX 4080 GPU & RTX 4080 GPU \\
    \midrule
    Activation function & ReLU & - \\
    Conv layer 1 (Channels, Kernel) & [1→32, 5×5] & - \\
    MaxPool 1 (Kernel, Stride) & 2×2, 2 & - \\
    Conv layer 2 (Channels, Kernel) & [32→16, 5×5] & - \\
    MaxPool 2 (Kernel, Stride) & 2×2, 2 & - \\
    Flattened size & 1248 & - \\
    FC layer 1 (Input, Output) & [1248, 128] & - \\
    FC layer 2 (Input, Output) & [128, 12] & - \\
    \midrule
    LSTM layer (Input, Hidden, Layers) & - & [104, 64, 2] \\
    FC layer (Input, Output) & - & [64,12] \\
    \bottomrule
    \end{tabular}
    }
    \label{tab-trainparams}
\end{table}

\subsection{Classification Results and Discussion}
\begin{figure}[t]
\centering
\includegraphics[width=\columnwidth]{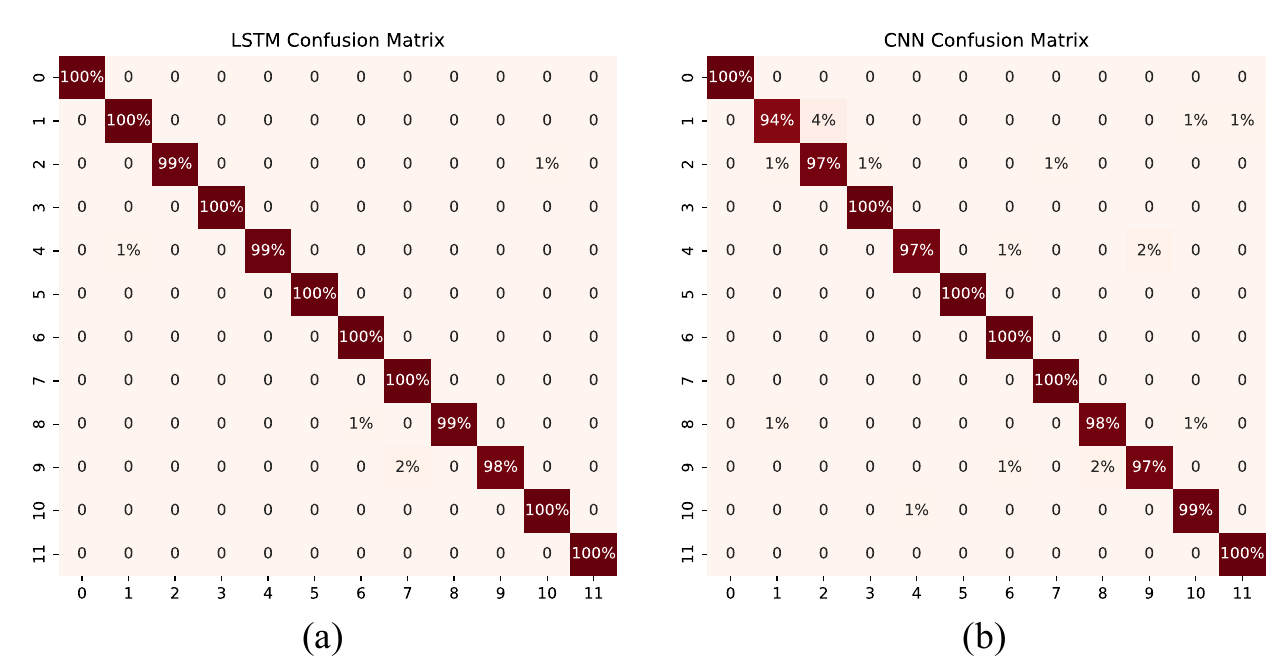}
\caption{Confusion matrices of tactile classification result. (a) The confusion matrix for the LSTM model. (b) The confusion matrix for the CNN model.}
\label{fig-classresult}
\end{figure}
As shown in the confusion matrices in Fig. \ref{fig-classresult}, the sensor system effectively distinguishes between 12 tactile modalities based on analysing 1200 samples. The CNN-based model achieves a classification accuracy of 98.5\%, while the LSTM-based model further improves the performance, reaching an accuracy of 99.6\%.  Table \ref{tab-classsota} presents a comparative summary with prior studies. While earlier works based on CNN architectures handled fewer touch categories (4 to 10 classes) with accuracies ranging from 93.3\% to 98.7\%, our method extends the classification to 12 categories without compromising accuracy. These results highlight the robustness and reliability of our approach in handling complex touch interaction scenarios.
\begin{table}[ht]
    \centering
    \caption{Comparison of various studies in classifying touch modalities}
    \label{tab-classsota}
    \begin{tabular}{l|c|c|c}
    \hline
    \textbf{Study} & \textbf{Number of Classes} & \textbf{Model} & \textbf{Accuracy} \\
    \hline
    Park et al. \cite{park2022biomimetic} & 4  & \multirow{3}{*}{CNN} & 98.7\% \\
    Yang et al. \cite{yang2023robotic}     & 9  &                     & 93.3\% \\
    Yang et al. \cite{yang2024body}        & 10 &                     & 95.3\% \\
    \hline
    \multirow{2}{*}{\textbf{Ours}} & \multirow{2}{*}{12} & CNN  & 98.5\% \\
                                        &                      & LSTM & 99.6\% \\
    \hline
    \end{tabular}
\end{table}

\section*{Conclusion}
We developed a lattice-structured flexible EIT tactile sensor, systematically optimized via 3D coupling field simulations and validated through real-world experiments. The sensor demonstrates high sensitivity, spatial resolution, and durability, achieving a CC of 0.9275, PSNR of 29.03 dB, SSIM of 0.9660, and RE as low as 0.3798, in reconstructing various touch patterns. It also accurately classifies 12 tactile gestures with over 99.6\% accuracy, confirming its effectiveness for both tactile reconstruction and interpretation. These results highlight its strong potential for HMI and wearable applications. Future work will focus on scaling the design and integrating multi-modal sensing to support next-generation electronic skin systems.

\bibliographystyle{IEEEtran}
\bibliography{References}

\begin{thebibliography}{10}
\providecommand{\url}[1]{#1}
\csname url@samestyle\endcsname
\providecommand{\newblock}{\relax}
\providecommand{\bibinfo}[2]{#2}
\providecommand{\BIBentrySTDinterwordspacing}{\spaceskip=0pt\relax}
\providecommand{\BIBentryALTinterwordstretchfactor}{4}
\providecommand{\BIBentryALTinterwordspacing}{\spaceskip=\fontdimen2\font plus
\BIBentryALTinterwordstretchfactor\fontdimen3\font minus \fontdimen4\font\relax}
\providecommand{\BIBforeignlanguage}[2]{{%
\expandafter\ifx\csname l@#1\endcsname\relax
\typeout{** WARNING: IEEEtran.bst: No hyphenation pattern has been}%
\typeout{** loaded for the language `#1'. Using the pattern for}%
\typeout{** the default language instead.}%
\else
\language=\csname l@#1\endcsname
\fi
#2}}
\providecommand{\BIBdecl}{\relax}
\BIBdecl

\bibitem{booth2018omniskins}
J.~W. Booth, D.~Shah, J.~C. Case, E.~L. White, M.~C. Yuen, O.~Cyr-Choiniere, and R.~Kramer-Bottiglio, ``Omniskins: Robotic skins that turn inanimate objects into multifunctional robots,'' \emph{Science Robotics}, vol.~3, no.~22, p. eaat1853, 2018.

\bibitem{li2022multifunctional}
G.~Li, S.~Liu, Q.~Mao, and R.~Zhu, ``Multifunctional electronic skins enable robots to safely and dexterously interact with human,'' \emph{Advanced Science}, vol.~9, no.~11, p. 2104969, 2022.

\bibitem{dai2024self}
W.~Dai, M.~Lei, Z.~Dai, S.~Ding, F.~Wang, D.~Fang, R.~Wang, B.~Qi, G.~Zhang, and B.~Zhou, ``Self-adhesive electronic skin with bio-inspired 3d architecture for mechanical stimuli monitoring and human-machine interactions,'' \emph{Small}, p. 2406564, 2024.

\bibitem{lai2023emerging}
Q.-T. Lai, X.-H. Zhao, Q.-J. Sun, Z.~Tang, X.-G. Tang, and V.~A. Roy, ``Emerging mxene-based flexible tactile sensors for health monitoring and haptic perception,'' \emph{Small}, vol.~19, no.~27, p. 2300283, 2023.

\bibitem{lu2023wearable}
D.~Lu, T.~Liu, X.~Meng, B.~Luo, J.~Yuan, Y.~Liu, S.~Zhang, C.~Cai, C.~Gao, J.~Wang \emph{et~al.}, ``Wearable triboelectric visual sensors for tactile perception,'' \emph{Advanced Materials}, vol.~35, no.~7, p. 2209117, 2023.

\bibitem{pohtongkam2021tactile}
S.~Pohtongkam and J.~Srinonchat, ``Tactile object recognition for humanoid robots using new designed piezoresistive tactile sensor and dcnn,'' \emph{Sensors}, vol.~21, no.~18, p. 6024, 2021.

\bibitem{niu2022ultrasensitive}
H.~Niu, Y.~Chen, E.-S. Kim, W.~Zhou, Y.~Li, and N.-Y. Kim, ``Ultrasensitive capacitive tactile sensor with heterostructured active layers for tiny signal perception,'' \emph{Chemical Engineering Journal}, vol. 450, p. 138258, 2022.

\bibitem{funk2024evetac}
N.~Funk, E.~Helmut, G.~Chalvatzaki, R.~Calandra, and J.~Peters, ``Evetac: An event-based optical tactile sensor for robotic manipulation,'' \emph{IEEE Transactions on Robotics}, 2024.

\bibitem{hu2022wireless}
H.~Hu, C.~Zhang, C.~Pan, H.~Dai, H.~Sun, Y.~Pan, X.~Lai, C.~Lyu, D.~Tang, J.~Fu \emph{et~al.}, ``Wireless flexible magnetic tactile sensor with super-resolution in large-areas,'' \emph{ACS nano}, vol.~16, no.~11, pp. 19\,271--19\,280, 2022.

\bibitem{wu2024bimodal}
Q.~Wu, C.~Zhou, Y.~Xu, S.~Han, A.~Chen, J.~Zhang, Y.~Chen, X.~Yang, J.~Huang, and L.~Guan, ``Bimodal intelligent electronic skin based on proximity and tactile interaction for pressure and configuration perception,'' \emph{ACS sensors}, vol.~9, no.~4, pp. 2091--2100, 2024.

\bibitem{wang2023tactile}
C.~Wang, C.~Liu, F.~Shang, S.~Niu, L.~Ke, N.~Zhang, B.~Ma, R.~Li, X.~Sun, and S.~Zhang, ``Tactile sensing technology in bionic skin: A review,'' \emph{Biosensors and Bioelectronics}, vol. 220, p. 114882, 2023.

\bibitem{van2020large}
L.~Van~Duong \emph{et~al.}, ``Large-scale vision-based tactile sensing for robot links: Design, modeling, and evaluation,'' \emph{IEEE Transactions on Robotics}, vol.~37, no.~2, pp. 390--403, 2020.

\bibitem{chen2024correcting}
H.~Chen, X.~Yang, G.~Ma, and X.~Wang, ``Correcting non-uniform sensitivity in eit tactile sensing via jacobian vector approximation,'' \emph{IEEE Robotics and Automation Letters}, 2024.

\bibitem{dong2024learning}
H.~Dong, X.~Wu, D.~Hu, Z.~Liu, F.~Giorgio-Serchi, and Y.~Yang, ``Learning-enhanced electronic skin for tactile sensing on deformable surface based on electrical impedance tomography,'' \emph{IEEE Transactions on Instrumentation and Measurement}, 2024.

\bibitem{wu2022new}
H.~Wu, B.~Zheng, H.~Wang, and J.~Ye, ``New flexible tactile sensor based on electrical impedance tomography,'' \emph{Micromachines}, vol.~13, no.~2, p. 185, 2022.

\bibitem{kim2024extremely}
K.~Kim, J.-H. Hong, K.~Bae, K.~Lee, D.~J. Lee, J.~Park, H.~Zhang, M.~Sang, J.~E. Ju, Y.~U. Cho \emph{et~al.}, ``Extremely durable electrical impedance tomography--based soft and ultrathin wearable e-skin for three-dimensional tactile interfaces,'' \emph{Science Advances}, vol.~10, no.~38, p. eadr1099, 2024.

\bibitem{chen2022large}
H.~Chen, X.~Yang, P.~Wang, J.~Geng, G.~Ma, and X.~Wang, ``A large-area flexible tactile sensor for multi-touch and force detection using electrical impedance tomography,'' \emph{IEEE Sensors Journal}, vol.~22, no.~7, pp. 7119--7129, 2022.

\bibitem{park2021deep}
H.~Park, K.~Park, S.~Mo, and J.~Kim, ``Deep neural network based electrical impedance tomographic sensing methodology for large-area robotic tactile sensing,'' \emph{IEEE Transactions on Robotics}, vol.~37, no.~5, pp. 1570--1583, 2021.

\bibitem{jamshidi2024design}
M.~Jamshidi, C.~B. Park, and F.~Azhari, ``The design and fabrication of a wearable lattice-patterned 3d sensing skin,'' \emph{Sensors and Actuators A: Physical}, vol. 369, p. 115143, 2024.

\bibitem{park2022biomimetic}
K.~Park, H.~Yuk, M.~Yang, J.~Cho, H.~Lee, and J.~Kim, ``A biomimetic elastomeric robot skin using electrical impedance and acoustic tomography for tactile sensing,'' \emph{Science Robotics}, vol.~7, no.~67, p. eabm7187, 2022.

\bibitem{yang2023robotic}
M.~J. Yang, K.~Park, W.~D. Kim, and J.~Kim, ``Robotic skin mimicking human skin layer and pacinian corpuscle for social interaction,'' \emph{IEEE/ASME Transactions on Mechatronics}, 2023.

\bibitem{yang2024body}
M.~J. Yang, H.~Chung, Y.~Kim, K.~Park, and J.~Kim, ``A body-scale robotic skin using distributed multimodal sensing modules: Design, evaluation, and application,'' \emph{IEEE Transactions on Robotics}, 2024.

\bibitem{jamshidi2023eit}
M.~Jamshidi, C.~B. Park, and F.~Azhari, ``An eit-based piezoresistive sensing skin with a lattice structure,'' \emph{Materials \& Design}, vol. 233, p. 112227, 2023.

\bibitem{zhu20203d}
Z.~Zhu, H.~S. Park, and M.~C. McAlpine, ``3d printed deformable sensors,'' \emph{Science advances}, vol.~6, no.~25, p. eaba5575, 2020.

\bibitem{Vauhkonen1998}
M.~Vauhkonen, D.~Vad{\'a}sz, P.~A. Karjalainen, E.~Somersalo, and J.~P. Kaipio, ``Tikhonov regularization and prior information in electrical impedance tomography,'' \emph{IEEE transactions on medical imaging}, vol.~17, no.~2, pp. 285--293, 1998.

\bibitem{dong2025data}
H.~Dong, R.~B. Liu, L.~Micklem, S.~P. E, F.~Giorgio-Serchi, and Y.~Yang, ``Data-efficient tactile sensing with electrical impedance tomography,'' \emph{IEEE Sensors Journal}, 2025.

\bibitem{wang2004image}
Z.~Wang, A.~C. Bovik, H.~R. Sheikh, and E.~P. Simoncelli, ``Image quality assessment: from error visibility to structural similarity,'' \emph{IEEE transactions on image processing}, vol.~13, no.~4, pp. 600--612, 2004.

\bibitem{yu2024high}
H.~Yu, H.~Liu, Z.~Liu, Z.~Wang, and J.~Jia, ``High-resolution conductivity reconstruction by electrical impedance tomography using structure-aware hybrid-fusion learning,'' \emph{Computer methods and programs in biomedicine}, vol. 243, p. 107861, 2024.

\end{thebibliography}

\vfill
\end{document}